\documentclass[journal,twoside,web]{ieeecolor}
\usepackage{generic}
\usepackage{cite}
\usepackage{amsmath,amssymb,amsfonts}
\usepackage{algorithmic}
\usepackage{booktabs}
\usepackage{graphicx}
\usepackage{algorithm,algorithmic}
\usepackage{hyperref}
\usepackage{url}
\usepackage{textcomp}
\usepackage{multirow}
\def\BibTeX{{\rm B\kern-.05em{\sc i\kern-.025em b}\kern-.08em
    T\kern-.1667em\lower.7ex\hbox{E}\kern-.125emX}}
\begin{document}
\title{Stain-aware Domain Alignment for Imbalance Blood Cell Classification}
\author{Yongcheng Li, Lingcong Cai, Ying Lu, Xianghua Fu, Xiao Han, Ma Li, Wenxing Lai, Xiangzhong Zhang, and Xiaomao Fan \IEEEmembership{Member, IEEE}
\thanks{This work is partially supported by the Special subject on Agriculture and Social Development, Key Research and Development Plan in Guangzhou (2023B03J0172), the Basic and Applied Basic Research Project of Guangdong Province (2022B1515130009), the Natural Science Foundation of Top Talent of SZTU (GDRC202318), the Stable Support Project for Shenzhen Higher Education Institutions (SZWD2021011), and the Research Promotion Project of Key Construction Discipline in Guangdong Province (2022ZDJS112).}
\thanks{Yongcheng Li, Lingcong Cai, Xianghua Fu, and Xiaomao Fan are with the College of Big Data and Internet, Shenzhen Technology University, Shenzhen 518118, China (e-mail: vincentleen30@gmail.com; cailingcong@gmail.com; fuxianghua@sztu.edu.cn; astrofan2008@gmail.com). }
\thanks{Ying Lu, Xiao Han, Wenxing Lai, and Xiangzhong Zhang are with the Department of Hematology, the Third Affiliated Hospital of Sun Yat-sen University,
Guangzhou, China (e-mail: luying5@mail.sysu.edu.cn; 973096161@qq.com; laiwx@mail.sysu.edu.cn; zhxzhong@mail.sysu.edu.cn).}
\thanks{Ma Li is with the Department of Clinical Laboratory, the Third Affiliated Hospital of Sun Yat-sen University, Guangzhou, China (e-mail: lmzslzc@163.com).}
\thanks{Yongcheng Li, Lingcong Cai, and Ying Lu contribute equally.}
\thanks{Xiaomao Fan and Xiangzhong Zhang are co-corresponding authors.}}

\maketitle

\begin{abstract}
% 150-250
Blood cell identification is critical for hematological analysis as it aids physicians in diagnosing various blood-related diseases. In real-world scenarios, blood cell image datasets often present the issues of domain shift and data imbalance, posing challenges for accurate blood cell identification. To address these issues, we propose a novel blood cell classification method termed SADA via stain-aware domain alignment. The primary objective of this work is to mine domain-invariant features in the presence of domain shifts and data imbalances. To accomplish this objective, we propose a stain-based augmentation approach and a local alignment constraint to learn domain-invariant features. Furthermore, we propose a domain-invariant supervised contrastive learning strategy to capture discriminative features. We decouple the training process into two stages of domain-invariant feature learning and classification training, alleviating the problem of data imbalance. Experiment results on four public blood cell datasets and a private real dataset collected from the Third Affiliated Hospital of Sun Yat-sen University demonstrate that SADA can achieve a new state-of-the-art baseline, which is superior to the existing cutting-edge methods with a big margin. The source code can be available at the URL (\url{https://github.com/AnoK3111/SADA}).

% collected from the Third Affiliated Hospital of Sun Yat-sen University
\end{abstract}


\section{Introduction}
\label{section 1}
\begin{figure}[tb]
    \centering
    \includegraphics[width=1.0\linewidth]{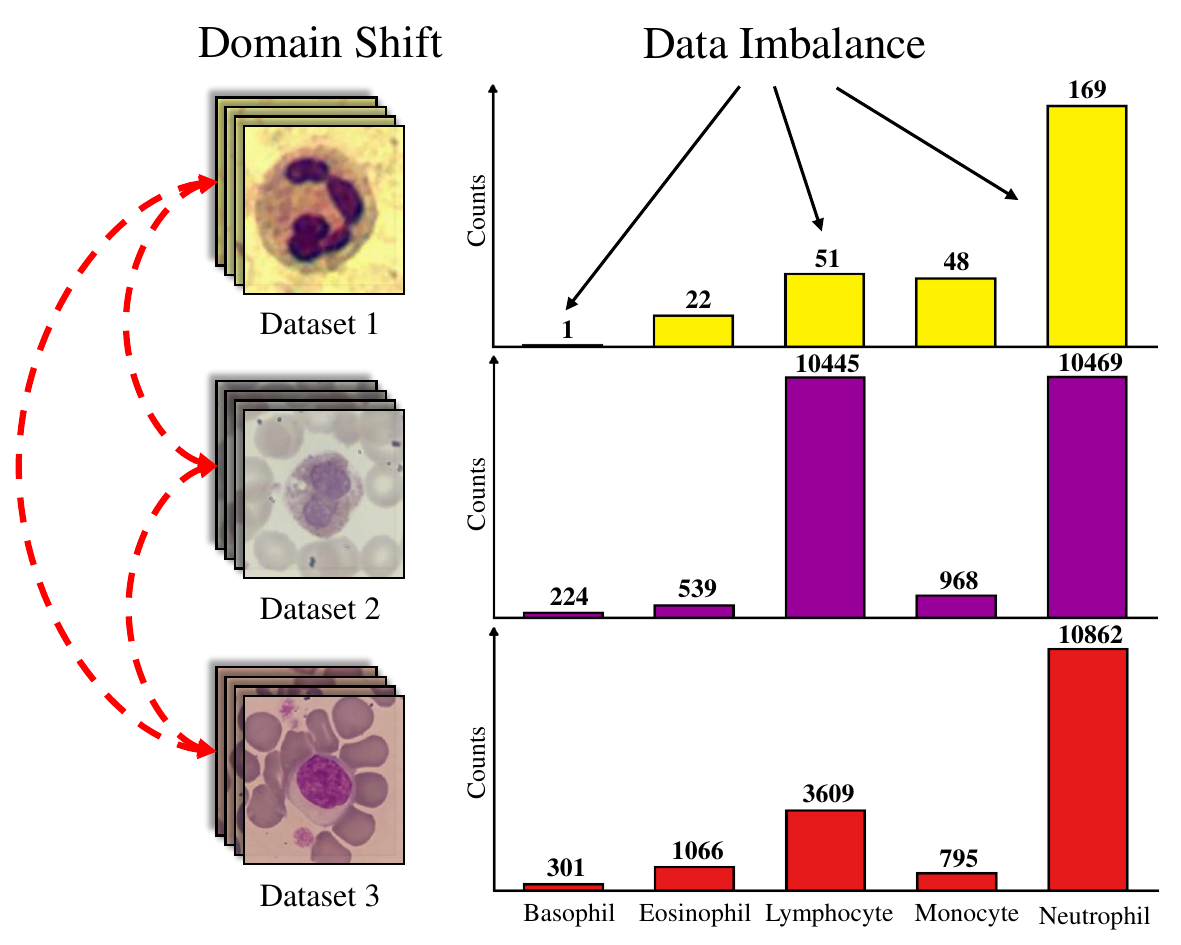}
    \caption{Blood cell image datasets often exhibit significant domain shifts and data imbalance, resulting in suboptimal performance in classification.}
    \label{intro}
\end{figure} 
Blood cell analysis serves as a prevalent diagnostic method for a spectrum of diseases, assisting physicians in identifying and monitoring various medical conditions such as leukemia, anemia, infections, autoimmune disorders, and other blood-related diseases \cite{wang2021research}. Therefore, the accurate and prompt classification of blood cells is crucial in clinical practice, playing a pivotal role in the timely detection of blood cell disorders. This process is at the forefront of medical diagnostics, ensuring effective medical treatment through early recognition and management of blood-related ailments.

In clinical practice, the identification of blood cells traditionally involves manual examination by skilled experts using a microscope, which is a repetitive, labor-intensive, and time-consuming process. Recently, significant advancements have been made in deep learning models for solving real-world classification tasks in various areas \cite{liu2023revisiting}\cite{van2023pdisconet}. Some researchers also attempted to utilize deep learning models in blood cell classification \cite{umer2023imbalanced}\cite{hehr2023explainable}, demonstrating promising results. However, blood cell images collected from various laboratories and hospitals often exhibit significant domain shifts and data imbalances, which is shown in Fig. \ref{intro}. The domain shifts can lead to a deterioration in the model's generalization performance, resulting in suboptimal performance on unseen datasets, while the data imbalance can cause the model to exhibit poor performance of the minority classes.

Regarding the issue of data imbalance, the existing methods to address this issue can be mainly categorized into three groups: class re-balancing, multi-expert learning, and multi-stage training. Class re-balancing including re-sampling \cite{jamal2020rethinking} and re-weighting \cite{xu2022constructing}, designed to recalibrate the contribution of each class during training, enhances overall performance but usually at the sacrifice of the accuracy of the majority classes. Multi-expert learning \cite{cai2021ace}\cite{li2022nested} utilizes diverse models to acquire knowledge from different aspects, resulting in an augmented complexity during the inference phase. Multi-stage training \cite{kang2019decoupling}\cite{Kang_2021} segments the training process into multiple stages, which can achieve competitive performance without altering the model's architecture.

Regarding the issue of domain shift, domain generalization serves as an effective solution by extracting domain-invariant features from different domains. It can be implemented through diverse methods, encompassing data augmentation and contrastive learning. Data augmentation extends the training dataset, enabling the model to better adapt to diverse domain features, including Mixup \cite{xu2020adversarial} and AutoAugment \cite{Cubuk_2019}. However, these methods overlooked that blood cell images inherently contain domain-invariant features (i.e. the morphology of blood cells). Contrastive learning utilizes various constraints to optimize contrastive-based objectives so that the model can acquire generalized features by leveraging diverse sample-to-sample relationships across various domains, like PCL \cite{yao2022pcl} and SelfReg \cite{kim2021selfreg}. However, these methods utilize the raw samples with domain-specific information as anchors, leading to the failure of contrastive learning to some extent.

% contrastive-based objective

 % To sum up, there are still significant challenges that need to be solved: 1) how to 

In this paper, we propose a novel domain alignment method, called SADA, for imbalanced multi-domain blood cell classification via stain-aware domain alignment. Specifically, we first introduce a stain-based augmentation approach that generates domain-transformed samples containing domain-specific information from source domains while preserving domain-invariant features. To encourage the model to capture the domain-invariant features, we design a local alignment constraint to make raw sample and domain-transformed samples consistent at the feature map level. Furthermore, we propose a domain-invariant supervised contrastive learning strategy by averaging raw sample and domain-transformed samples as anchors, which can mitigate the impact of domain-specific features and facilitate discriminative feature learning. It is worth noting that we decouple the training process into domain-invariant feature learning and classification training, alleviating the problem of data imbalance. Experimental results on four public blood cell datasets and a real dataset obtained from the Third Affiliated Hospital of Sun Yat-sen University named SYSU3H (IRB No. RG2023-265-01) demonstrate that our proposed SADA achieves the state-of-the-art results with a significant margin. Our main contributions can be summarized as follows:
% obtained from the Third Affiliated Hospital of Sun Yat-sen University
\begin{enumerate}
    \item We propose a stain-based augmentation approach to generate domain-transformed samples, enriching sample domain diversity as well as further enhancing the sample volume.
    \item We design a local alignment constraint to make the target sample and generated samples consistent at the feature map level, which encourages the model to capture the domain-invariant features.    
    \item We propose a domain-invariant supervised contrastive learning strategy by averaging raw sample and domain-transformed samples as anchors, which enables the model to learn more discriminative features.
    \item We decouple the training process into two stages of domain-invariant feature learning and classification training, which can alleviate the problem of data imbalance.
    \item Extensive experiment results conducted on four public blood cell datasets and a private blood cell dataset show that our proposed SADA can achieve the state-of-the-art results with a big margin. 
\end{enumerate}

The remainder of this paper is structured as follows. Section \ref{rw} reviews the related work on data imbalance and domain generalization. Section \ref{section:3} presents the details of the proposed SADA. The experimental results are discussed in Section \ref{section 4.1}, followed by the conclusion of this paper in Section \ref{con}.

\section{Related Work}
\label{rw}

\subsection{Data Imbalance}
\label{section:2.3}

\noindent
\textbf{Class re-balancing.} Class re-balancing strategies, encompassing class re-sampling\cite{cao2019learning}\cite{jamal2020rethinking} and loss re-weighting \cite{lin2017focal}\cite{xu2022constructing}, are employed to rectify the imbalanced contributions of individual classes. By adjusting the number of instances in each class, class re-sampling ensures that the model receives a more balanced set of examples, preventing it from being biased towards the majority classes. Regarding Loss re-weighting addresses data imbalance by adjusting the loss contribution of each class. This ensures that minority classes have a proportionally greater impact on the training process, leading to a more balanced and robust model. These methods are designed to emphasize minority classes, albeit with the potential trade-off of compromising the classification accuracy of majority classes.

\noindent
\textbf{Multi-expert learning.} Multi-expert learning \cite{li2020overcoming}\cite{cai2021ace}\cite{li2022nested} leverages multiple specialized models, each of which is trained on a specific subset of the data, each benefiting from the specialization in their dominant domain. This approach can effectively mitigate the challenges of imbalanced datasets by ensuring that each class is adequately represented and learned by at least one expert. For instance, MDCS \cite{zhao2023mdcs} trained various experts on different parts of the dataset with self-distillation for imbalanced data classification. This technique offers a flexible and effective solution to the challenges posed by imbalanced datasets. However, this can lead to an increase in the number of parameters and the overall complexity of the model.

\noindent
\textbf{Multi-stage training.} The multi-stage training \cite{Kang_2021}\cite{kang2019decoupling}\cite{wang2021contrastive} addresses data imbalance by dividing the training process into distinct stages, enabling focused learning on minority classes and improving overall performance. This approach encourages the model to learn the important features from both majority and minority classes, thereby mitigating the bias towards the majority class. For instance, Kang et al.\cite {kang2019decoupling} decoupled the training process into representation learning and classification training to alleviate the impact of data imbalance in the stage of representation learning. Its most direct advantage lies in enabling the model to allocate more resources to identify minority classes without altering the model architecture. Therefore, this method has been applied in our study.

\subsection{Domain Generalization}

\noindent
\textbf{Augmentation.} Existing augmentation methods primarily focus on two dimensions: feature-level enhancement and image-level augmentation. Feature-level enhancement \cite{li2021simple}\cite{nam2021reducing}\cite{liu2024cross} involves the combination of distinct features or their stylistic statistics. Image-level augmentation \cite{chang2021stain}\cite{kang2022style} \cite{shen2022randstainna} encompasses direct image blending or the synthesis of new data through phase concatenation. By exposing the model to a wider range of scenarios through feature-level and image-level augmentations, encourages it to extract domain invariant features, thereby improving its generalization performance. However, most of the current augmentation methods compromise the domain-invariant features of the blood cell images, undermining the models' generalization abilities.

\noindent
\textbf{Contrastive learning.} Most studies\cite{umer2023imbalanced}\cite{tan2024rethinking} in domain generalization concentrate on extracting domain-invariant representations. A prevalent approach is to minimize statistical metrics, such as Maximum Mean Discrepancy (MMD)\cite{li2018domain}. Additionally, inspired by recent progress in supervised contrastive learning\cite{khosla2020supervised}\cite{wang2020learning}, some works\cite{kim2021selfreg}\cite{jeon2021feature} managed to minimize the distance between samples of the same class (positive pair) in the representation space while others (negative pair) far apart. For instance, PDEN\cite{li2021progressive} utilizes contrastive learning to address single-domain generalization. SelfReg\cite{kim2021selfreg} further explores the alignment of positive pairs through a self-contrastive approach. This scheme has been widely applied in the field of domain generalization, achieving promising results. However, most contrastive learning methods often ignore the issue that the selected anchors may contain domain-specific features, leading to the failure of contrastive learning to some extent.

\renewcommand{\dblfloatpagefraction}{.7}
\begin{figure*}[tb]
    \centering
    \includegraphics[width=0.95\linewidth]{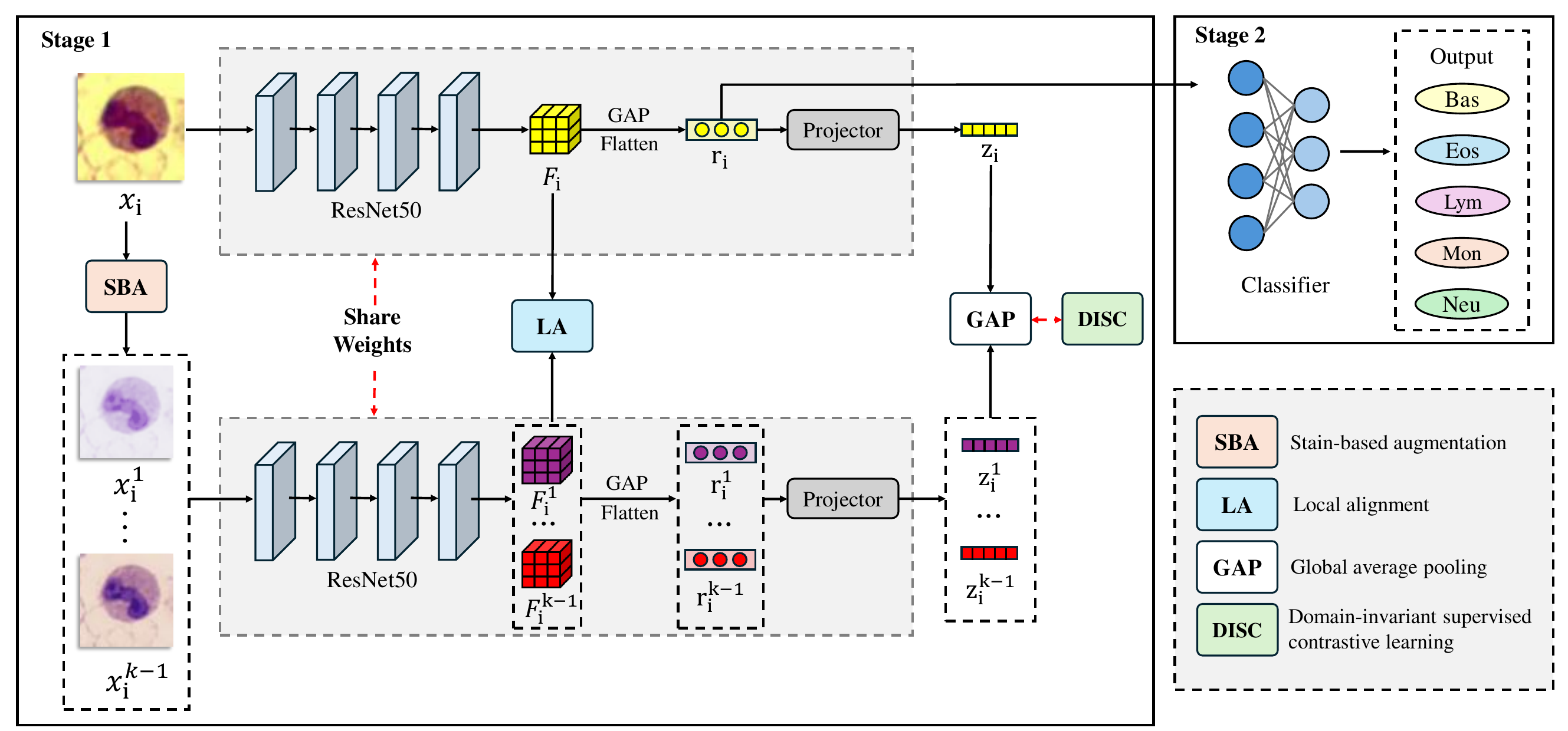}
    \caption{The pipeline of the proposed SADA. It is composed of two stages: domain-invariant feature learning and blood cell classification. In stage 1, we first leverage the stain-based augmentation module to generate domain-transformed samples. To enable the model to learn domain-invariant features, the corresponding generated feature maps are fed into the local alignment module after passing through the backbone network. Finally, we leverage the domain-invariant supervised contrastive learning for further domain-invariant feature learning. In stage 2, we leverage a fully connected layer for subsequent blood cell classification.}
    \label{overview}
\end{figure*}

\section{Methodology}
\label{section:3}
Fig. \ref{overview} shows the overall framework of the proposed SADA, including two stages: (1) domain-invariant feature learning and (2) blood cell classification. In stage 1, we first introduce the stain-based augmentation approach to generate domain-transformed samples, incorporating the different stain colors from the source domains (see Section \ref{section:3.1}).  Additionally, we propose a local alignment constraint to perform pixel-wise alignment, thereby encouraging the model to capture the domain-invariant feature (see Section \ref{section 3.2}). Moreover, we propose domain-invariant supervised contrastive learning to learn more discriminative features (see Section \ref{section 3.3}). It is worth noting that we decouple the training process into the domain-invariant feature learning and blood cell classification, mitigating the impact of data imbalance (see Section \ref{section 3.4}).

Formally, given the unseen target domain \(D_t\) that samples blood cell images from a distinct domain, the primary objective is to train a model on the multi-source \(D_s\) to perform well on the unseen target domain \(D_t\). Specially, let \(D_s = \{D_1, D_2, ..., D_M\}\)(with \(M > 1\)) represents a set of training domains, where \(D_s\) denotes a joint distribution over the blood cell images \(x\) and their corresponding labels \(y\). Finally, let \(f_{SADA}= (\cdot)\) denote our model for blood cell classification and the outputting corresponding to the blood cell label \(\hat{y}\).

\subsection{Stain-Based Augmentation}
\label{section:3.1}
\renewcommand{\dblfloatpagefraction}{.7}
\begin{figure}[tb]
    \centering
    \includegraphics[width=1.0\linewidth]{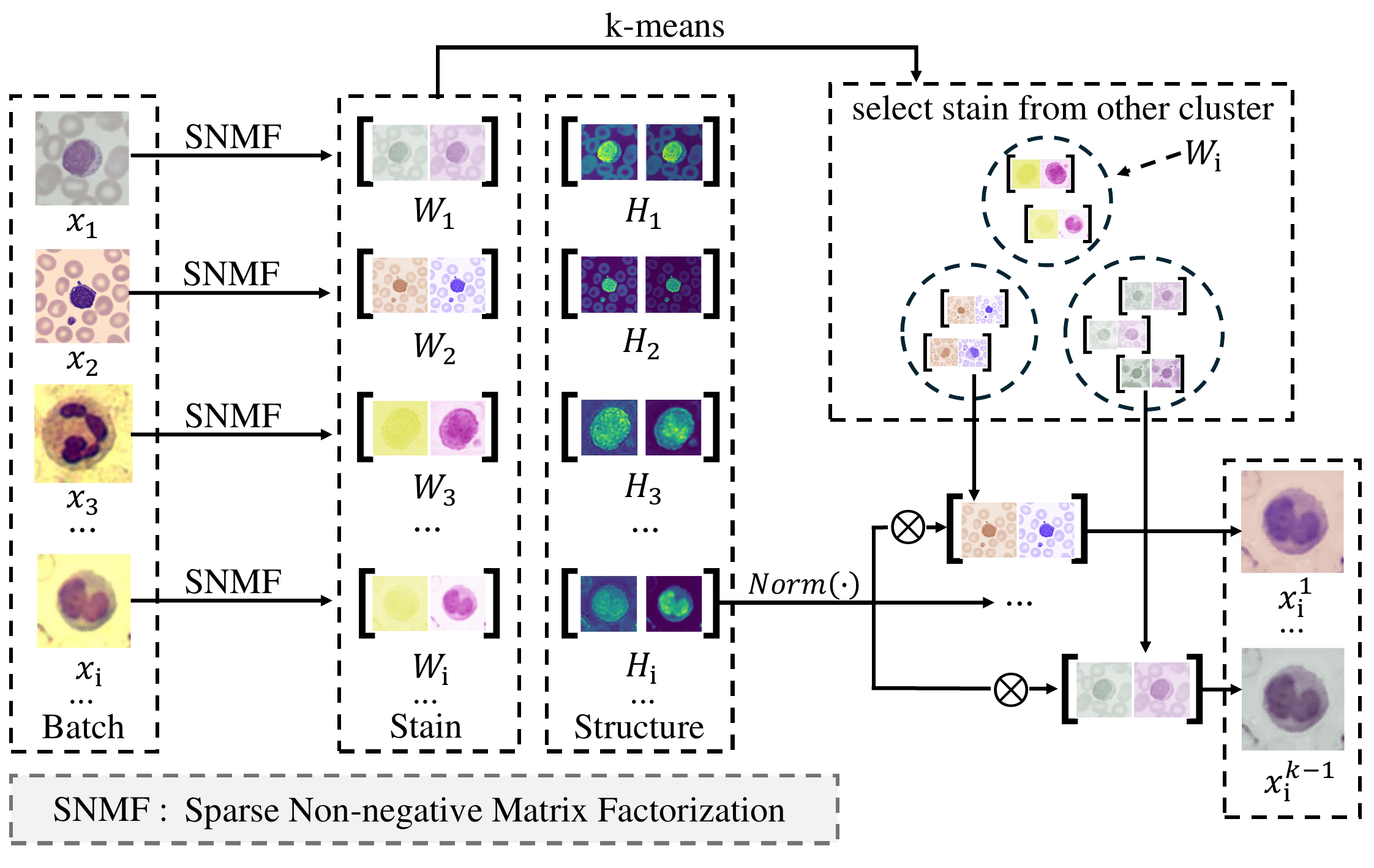}
    \caption{The overview of stain-based augmentation.}
    \label{fig:sba}
\end{figure} 
It is observed that generating domain-transformed samples incorporating different stain colors from training data can be helpful for the model to capture the domain-invariant feature of blood cell images. Building upon this insight, we introduce stain-based augmentation, a method designed to generate domain-transformed samples that incorporate diverse stain colors from different source domains without compromising the domain-invariant features. Fig. \ref{fig:sba} illustrates the general process, comprising three steps: blood cell image decomposition, stain color clustering, and domain-transformed sample generation.

\textbf{Blood cell image decomposition.} Staining solutions are employed for coloring blood cell images to enhance the visibility of cellular structures. The resulting color effect is contingent upon the specific type and quantity of staining solution absorbed by the blood cells. Therefore, we can decompose the blood cell images into density maps \(H_s\) (i.e. stain color) and stain color matrices \(W_s\) (i.e. structure) via the Beer-Lambert law:

\begin{equation}
\label{eq:1}
V=-\log_{}{\frac{X}{X_0}}=WH
\end{equation}
% where \(X \in \mathbb{R}^{3 \times n}\) represents the original blood cell image in the RGB space with \(n\) being the number of pixels, \(X_0\) is the illuminating light intensity on the sample (here \(X_0=255\) for 8-bit blood cell images), \(W \in \mathbb{R}^{3 \times r}\) is the stain color appearance matrix, where each column is a normalized color basis, and \(r\) is the number of bases. \(H \in \mathbb{R}^{r \times n}\) represents the density maps, describing the structure of the images \(X\). For each blood cell image \(X\), the Beer-Lambert transformation (BLT) in Eq. \ref{eq:1} is employed to obtain the optical density \(V \in \mathbb{R}^{3 \times n}\). Following the \cite{vahadane2016structure} work, we leverage sparse non-negative matrix Factorization (SNMF) methods to obtain the stain color matrices \(W\) and density maps \(H\).
\noindent
where \(X\) denotes blood cell image in RGB space (\(X \in \mathbb{R}^{3 \times n}\)) with \(n\) pixels. \(X_0\) is the illuminating light intensity (255 for 8-bit images). Stain color appearance matrix \(W \in \mathbb{R}^{3 \times r}\) contains normalized color bases, and \(H \in \mathbb{R}^{r \times n}\) represents density maps. Beer-Lambert transformation yields optical density \(V \in \mathbb{R}^{3 \times n}\). Following the \cite{vahadane2016structure} work, we leverage sparse non-negative matrix factorization (SNMF) methods to optimize the stain color matrices \(W\) and density maps \(H\).

\textbf{Stain color clustering.} After decomposing the blood cell images, we can acquire stain color matrices \(W_s\) and density maps \(H_s\) in a mini-batch. We can choose colors based on different domains, followed by the re-staining operation on the corresponding optical density, to generate domain-transformed samples. However, because of operational differences, even stain colors of blood cells from the same dataset can exhibit significant variations. Therefore, we employ a clustering approach to reassign domain labels based on the stain color matrices within a mini-batch. In this study, we utilize the k-means algorithm to cluster stain color matrices \(W_s\) in each mini-batch. Each cluster generated by k-means is assigned unique domain labels (\(d_i \in \{1,2, \ldots, k\}\)) for the samples within that cluster. The hyperparameter \(k\) defines the number of clusters in the k-means algorithm.

\textbf{Domain-transformed sample generation.} We randomly select stain color matrices \(W_t\) from other $k-1$ clusters to perform re-staining operations on the density maps \(H_i\) of the raw samples. Because of the scale difference of density maps, an additional normalization procedure is required for \(H_i\), 
\begin{equation}
\label{eq:5}
Norm(H_i,H_t) = H_i \cdot \frac{P(H_t)}{P(H_i)},
\end{equation}
where $P(\cdot)$ calculates the $99^{\text{th}}$ percentile for each row in density maps \(H_i\) and \(H_t\). After normalization, the domain-transformed samples \(x_i^*\) can be generated through the inverse Beer-Lambert transformation,
\begin{equation}
\label{eq:6}
x_i^* = X_0 \exp(-W_tNorm(H_i,H_t))
\end{equation}
\noindent
where $i \in \{1, 2, \cdots, N\}$ and $N$ is the size of the mini-batch of blood cell images.

\subsection{Local Alignment}
\label{section 3.2}

\begin{figure}[tb]
    \centering
    \includegraphics[width=1.0\linewidth]{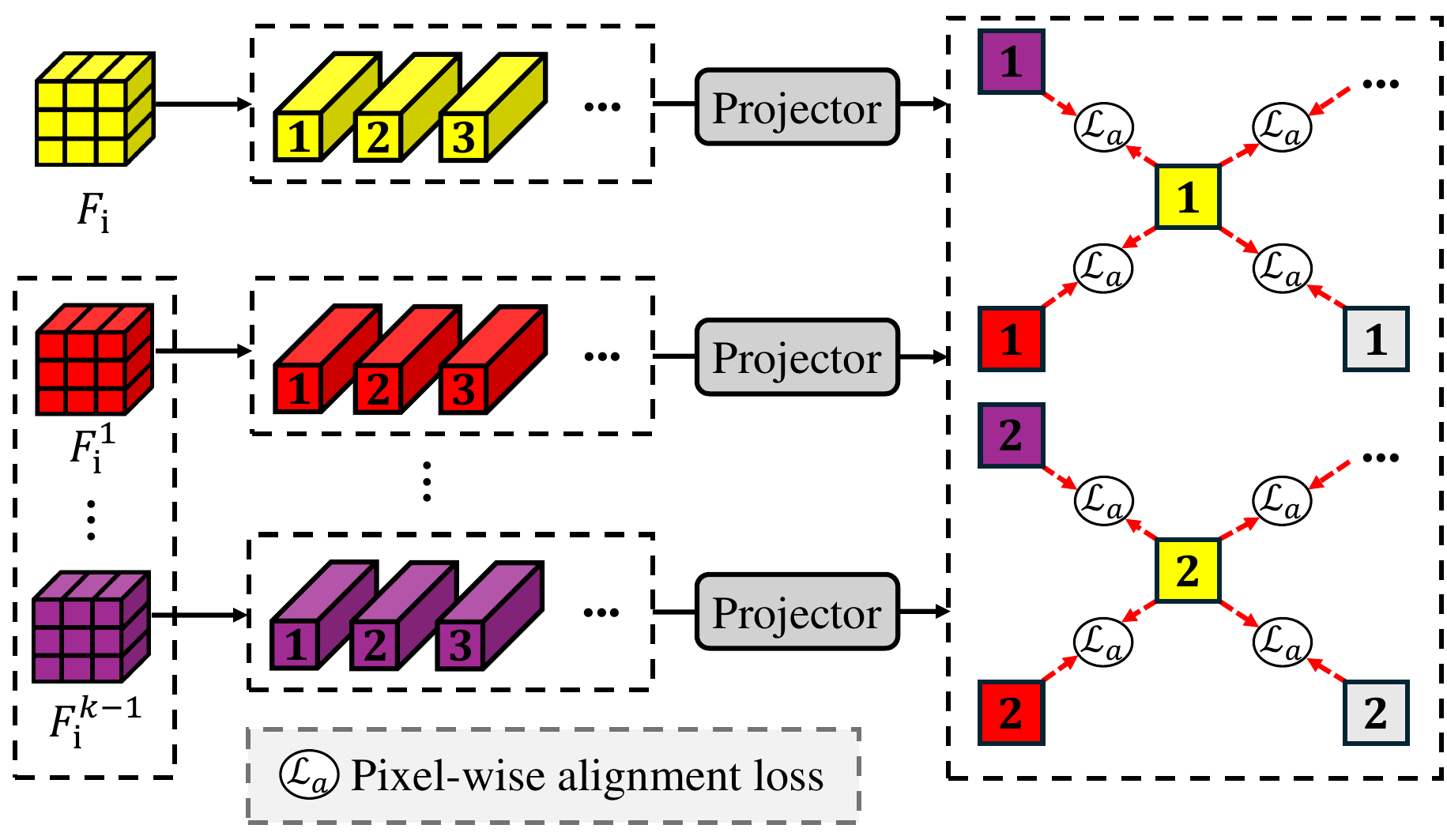}
    \caption{The overview of local alignment.}
    \label{fig:la}
\end{figure} 

As stain-based augmentation solely preserves the structure of the blood cell image, we employ the local alignment constraint to learn domain-invariant features between raw sample and domain-transformed samples at the feature map level (see Fig. \ref{fig:la}). Formally, let \(F_{i,p}, F_{i,p}^t \in \mathbb{R}^{D_b}\) represent the feature maps extracted from original and domain-transformed samples by the final layer of the backbone network (before the global average pooling layer), where \(D_b\) denotes the output channel of the backbone network, and \(p\) is the spatial location index of feature maps. To prevent performance degradation due to representation collapse, we utilize a non-linear multiple-layer perception (MLP) layer \(f_l(\cdot)\) to map \(F_{i,p}, F_{i,p}^t\) to the feature embeddings \(E_{i,p}, E_{i,p}^t \in \mathbb{R}^{D_e}\). With the feature embeddings \(E_{i,p}, E_{i,p}^t\), the pixel-wise alignment loss is calculated as follows:
\begin{equation}
\label{eq4}
\mathcal{L}_{a}(E_{i,p}) = \frac{1}{k-1} \sum_{t=1}^{k-1}\Vert \frac{E_{i,p}}{\Vert E_{i,p} \Vert_2} - \frac{E_{i,p}^t}{\Vert E_{i,p}^t \Vert_2} \Vert_2^2
\end{equation}
Referring to the pixel-wise alignment loss \(\mathcal{L}_{a}\) in Eq. (\ref{eq4}), the local alignment loss function can be mathematically formulated as:
\begin{equation}
\label{eq5}
\mathcal{L}_{LA} = \sum_{i=1}^{N} \sum_{p} \mathcal{L}_{a}(E_{i,p})
\end{equation}
\noindent
where \(N\) denotes the size of the mini-batch.

\subsection{Domain-Invariant Supervised Contrastive Learning}
\label{section 3.3}

\begin{figure}[tb]
    \centering
    \includegraphics[width=1.0\linewidth]{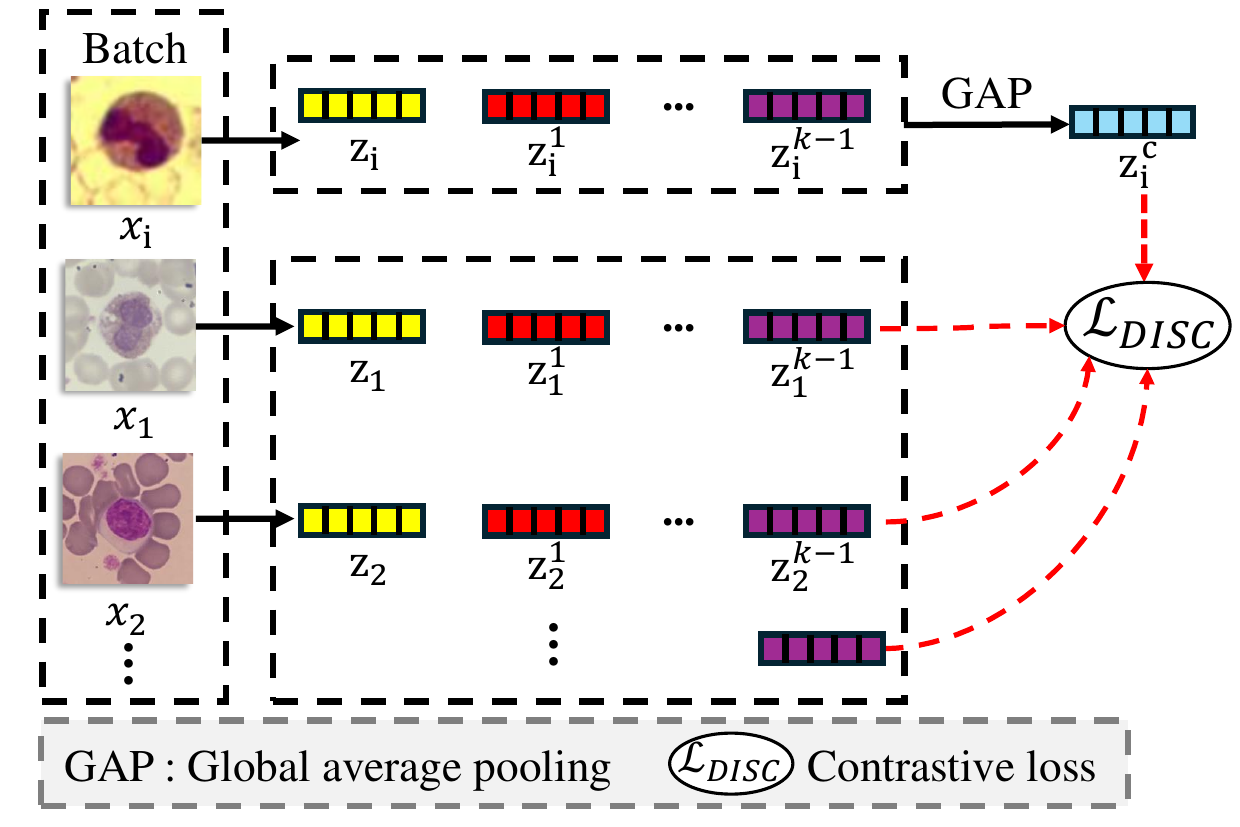}
    \caption{The overview of domain-invariant contrastive learning.}
    \label{fig:disc}
\end{figure} 

In the presence of domain shift, the selected anchors may include domain-specific features, leading to the failure of contrastive learning to some extent. Therefore, we calculate the averaging features of a raw sample and domain-transformed samples as anchors to obtain reference points that are less influenced by domain-specific features (see Fig. \ref{fig:disc}). Specifically, for each mini-batch, after undergoing the backbone \(f_{b}(\cdot)\) (i.e. ResNet50) and projector \(f_{p}(\cdot)\) (i.e. a linear layer), both the raw sample \(x_i\) and domain-transformed samples $\{x_i^1, x_i^2, \cdots, x_i^{k-1}\}$ obtain their corresponding feature embeddings $\{z_i, z_i^1, z_i^2, \cdots, z_i^{k-1}\}$, where \(z_i\) denotes the feature of raw sample. For each feature embedding $z$, L2 normalization is applied to utilize the dot product as a metric for distance. The anchor $a_i$ can be calculated by averaging the feature embeddings $\{z_i, z_i^1, z_i^2, \cdots, z_i^{k-1}\}$ among raw sample and domain-transformed samples. We define the positives for an anchor \(a_i\) as \(\{z_i^{+}\}=\{z_j|y_j=y_i\} \cup \{z_j^t|y_j=y_i\}\), where \(y_i\) is the class label of \(x_i\). Our proposed domain-invariant supervised contrastive loss function can be expressed as:
\begin{equation}
\label{eq7}
\mathcal{L}_{DISC}=\sum_{i=1}^{N}\frac{-1}{|\{z_i^{+}\}|}\sum_{z_j \in \{z_i^{+}\}}\log\frac{\exp (a_i \cdot z_j / \tau)}{\sum_{k=1}^{N} \exp (a_i \cdot z_k / \tau)}
\end{equation}
\noindent
where $\tau \in R^{+}$ is a scalar temperature parameter.

\subsection{Objective Loss}
\label{section 3.4}
Our training process is divided into two stages domain-invariant feature learning and blood cell classification. This division allows the model to concentrate more on feature domain-invariant feature learning, thereby alleviating the issue of data imbalance. Our domain-invariant feature learning loss function $\mathcal{L}_{r}$ can be formulated as follow,
% In the stage of representation learning, we utilize local alignment loss \(\mathcal{L}_{LA}\) in Eq.\ref{eq5} and domain-invariant supervised contrastive loss function \(\mathcal{L}_{DISC}\) in Eq. \ref{eq7}. In the stage of classifier learning,  we utilize cross-entropy loss function $\mathcal{L}_{CE}$ to train the classifier and backbone network. In this context, the classifier \(f_c(\cdot)\) is represented by a linear layer that projects the feature representation \(r\) into class-wise logits \(l = f_c(r) \in \mathbb{R}^{D_c}\).
\begin{equation}
\mathcal{L}_{r}=\mathcal{L}_{DISC}+\beta \mathcal{L}_{LA}
\end{equation}
where \(\beta\) denotes a trade-off hyperparameter between the two loss terms. The classifier learning loss function can be formulated by:
\begin{equation}
\mathcal{L}_{c} = \mathcal{L}_{CE}(y, \hat{y}) = -\sum_{i=1}^{C} y_i \log(\hat{y}_i)
\end{equation}
where \(C\) is the number of classes, \(y\) represents the one-hot encoded ground truth labels, and \(\hat{y}\) is the predicted probability distribution by the model.

\begin{table*}[tb]
  \begin{center} 
    \caption{Comparison with state-of-the-art methods on four public blood cell datasets. The best performance is in \textbf{bold} and the second best is indicated with \underline{underline}.}
    % Our proposed SADA outperforms all previous methods and achieves state-of-the-art performance.
    \label{result on four datasets}
    \resizebox{\textwidth}{!}{
    \begin{tabular}{l|cc|cc|cc|cc|cc}
    \hline
      \multirow{2}{*}{Methods}& \multicolumn{2}{c|}{\textbf{Acevedo-20}} & \multicolumn{2}{c|}{\textbf{LDWBC}} & \multicolumn{2}{c|}{\textbf{Raabin-WBC}} & \multicolumn{2}{c|}{\textbf{Zheng-18}} & \multicolumn{2}{c}{\textbf{Average}}\\
      \cline{2-11}
      & \textbf{F1-micro} & \textbf{F1-macro} & \textbf{F1-micro} & \textbf{F1-macro} & \textbf{F1-micro} & \textbf{F1-macro} & \textbf{F1-micro} & \textbf{F1-macro} & \textbf{F1-micro} & \textbf{F1-macro}\\
      \hline
      ERM\cite{vapnik1999overview} & 70.0$\pm$5.0 & 71.3$\pm$4.6 & 83.8$\pm$0.4 & 67.0$\pm$0.3 & 86.9$\pm$0.5 & 59.1$\pm$0.6 & 77.8$\pm$1.0 & 55.4$\pm$1.4 & 79.6$\pm$1.4 & 63.2$\pm$1.5 \\
      %MMD & 0 & 0 & 0 & 0 & 0 & 0\\
      CORAL\cite{sun2016deep} & \underline{70.7$\pm$3.8} & \underline{73.6$\pm$4.0} & 84.2$\pm$0.6 & 66.7$\pm$0.6 & \underline{87.2$\pm$0.2} & \underline{61.0$\pm$1.4} & 77.7$\pm$3.1 & 57.3$\pm$1.8 & 79.9$\pm$1.4 & \underline{64.7$\pm$1.2}\\
      Mixup\cite{xu2020adversarial} & 67.0$\pm$2.2 & 69.3$\pm$2.5 & 83.4$\pm$0.4 & 67.6$\pm$0.4 & \textbf{87.4$\pm$0.1} & \textbf{61.9$\pm$0.3} & 77.0$\pm$0.2 & 54.2$\pm$0.5 & 78.7$\pm$0.4 & 63.3$\pm$0.7\\
      SelfReg\cite{kim2021selfreg} & 68.4$\pm$1.9 & 70.3$\pm$1.4 & \underline{84.5$\pm$1.1} & \underline{67.8$\pm$1.3} & 86.8$\pm$0.2 & 60.0$\pm$0.2 & 76.5$\pm$2.2 & 53.4$\pm$3.3 & 79.0$\pm$0.4 & 62.9$\pm$0.7\\
      BoDA\cite{umer2023imbalanced} & 70.5$\pm$3.2 & \underline{73.6$\pm$3.3} & 82.2$\pm$0.5 & 65.8$\pm$0.6 & 86.8$\pm$0.0 & 60.6$\pm$0.5 & \underline{80.5$\pm$1.1} & \underline{58.7$\pm$2.6} & \underline{80.0$\pm$0.6} & \underline{64.7$\pm$0.2}\\
      Ours & \textbf{79.9}$\pm$\textbf{4.6} & \textbf{81.6}$\pm$\textbf{4.2} & \textbf{86.0}$\pm$\textbf{1.6} & \textbf{68.2}$\pm$\textbf{1.4} & \textbf{87.4}$\pm$\textbf{0.0} & \textbf{61.9}$\pm$\textbf{0.8} & \textbf{84.3}$\pm$\textbf{1.8} & \textbf{62.6}$\pm$\textbf{2.6} & \textbf{84.4}$\pm$\textbf{0.3} & \textbf{68.6}$\pm$\textbf{0.2}\\
      \hline
    \end{tabular}}
  \end{center}
\end{table*}
\section{Experiment Results and Discussion}
\subsection{Experimental Settings}
\label{section 4.1}
\noindent
\textbf{Datasets}. We perform extensive experiments on five blood cell image datasets: Acevedo-20\cite{Acevedo_2020}, Raabin-WBC\cite{kouzehkanan2022large}, LDWBC\cite{chen2021transmixnet}, Zheng\cite{zheng2018fast} and our private dataset termed SYSU3H (IRB No. RG2023-265-01) obtained from The Third Affiliated Hospital of Sun Yat-sen University. It is worth noting that the five datasets are collected from different laboratories and hospitals, which ensures the presence of domain shift among the different datasets.

\noindent
\textbf{Acevedo-20}: The dataset contains 17,092 blood cell images of normal individuals with 8 classes. The images with a resolution of \(360 \times 363\) pixels were generated by Sysmex SP1000i slide maker-stainer with May Grünwald-Giemsa staining.

\noindent
\textbf{Raabin-WBC}: It consists of 16,633 white blood cell images in 5 categories, captured with Olympus CX18 and Zeiss microscopes at 100x magnification. It contains images of varying dimensions in each category and all the samples are stained with Giemsa.

\noindent
\textbf{LDWBC}: It consists of 150 blood cell images from healthy individuals with 5 different categories, which were stained with the Wright\&Gimsa solution with the resolution of \(1280 \times 1280\).

\noindent
\textbf{Zheng-18}: The dataset includes 300 images of individual blood cells, cropped from 80 source images captured by a Motic Moticam Pro 252A optical microscope camera with a size of \(120 \times 120\) pixels.

\noindent
\textbf{SYSU3H}: Our dataset is obtained from The Third Affiliated Hospital of Sun Yat-sen University. It contains 331 blood cell images with 5 classes. All the samples are stained with Wright-Giemsa and captured by a Leica ICC50 HD digital microscope camera.

\noindent
\textbf{Evaluation metrics}. For improved model training and evaluation, we perform a five-class classification task involving neutrophils, eosinophils, basophils, monocytes, and lymphocytes. Due to dataset imbalance, we utilize the F1-macro score as an evaluation metric, reflecting the model's classification performance in handling data imbalance. Additionally, we report the F1-micro, offering a comprehensive metric through overall statistics across all categories. To ensure a fair comparison, all algorithms employ the same backbone network (i.e. ResNet50 pre-trained on ImageNet).

\noindent
\textbf{Implementation details}. Experiments are conducted on a computing server equipped with two NVIDIA GeForce RTX 4090 GPUs offering 48GB in total of video memory and an Intel(R) Core(TM) i9-10900X CPU providing 128GB of system memory. The training process uses Pytorch 2.1.1 in Python 3.9.12 with Adam optimizer and a learning rate of 0.00005. In addition, both the domain-invariant feature learning and classification training are trained for 4,000 steps respectively with a batch size of 32 for each of the training datasets. Following the training and model selection strategy in SWAD \cite{cha2021swad}, we utilize weight averaging to obtain a more robust model. 

% \renewcommand{\dblfloatpagefraction}{.7}
% \begin{figure*}[htbp]
%     \centering
%     \includegraphics[width=0.99\linewidth]{vis.pdf}
%     \caption{Visuallization.}
%     \label{overview}
% \end{figure*} 

% \begin{table*}[h!]
%   \begin{center} 
%     \caption{Comparison with state-of-the-art methods on a private real dataset. The best performance is in \textbf{bold} and the second best is indicated with \underline{underline}.}
%     % Our proposed SADA outperforms all previous methods and achieves state-of-the-art performance.
%     \label{table:3}
%     \resizebox{\textwidth}{!}{
%     \begin{tabular}{l|cccccc}
%       \hline
%      Methods & ERM\cite{vapnik1999overview} & CORAL\cite{sun2016deep} & Mixup\cite{xu2020adversarial} & 
%      SelfReg\cite{kim2021selfreg} & 
%      BoDA\cite{umer2023imbalanced} & SADA\\
%      \hline
%      F1-micro & 82.6$\pm$1.5 & \underline{83.3$\pm$0.6} & 79.8$\pm$2.2 & 80.8$\pm$1.2 & 79.8$\pm$0.9 & \textbf{85.4}$\pm$\textbf{0.3} \\
%      F1-macro & 83.0$\pm$1.3 & \underline{83.3$\pm$0.6} & 79.7$\pm$2.3 & 80.6$\pm$1.6 & 80.0$\pm$0.9 & \textbf{85.1}$\pm$\textbf{0.3}\\
%      \hline
%     \end{tabular}}
%   \end{center}
% \end{table*}

\subsection{Experimental Results}
\label{Section:4.2}

\noindent
\textbf{Results.} For all datasets, we adopt the leave-one-out strategy for evaluation, where one dataset serves as the hold-out target domain, and the remaining datasets act as source domains. Average results from the 3 trials of different compared methods and our method are listed in Table \ref{result on four datasets}. It is observed that our proposed SADA achieves an average F1-micro score of 84.4\% and an F1-macro score of 68.6\%, surpassing the second-best method, BoDA\cite{umer2023imbalanced}, by 4.4\% and 3.9\%, respectively.

For each dataset, our proposed method, SADA, consistently demonstrates superior performance across all evaluated blood cell datasets, achieving the highest F1-micro and F1-macro scores. Regarding the Acevedo-20 dataset, SADA achieved F1-micro and F1-macro scores of 79.9\% and 81.6\%, respectively, significantly outperforming the second-best method, CORAL\cite{sun2016deep}, which scored 70.7\% and 73.6\%. This improvement of 9.2\% in F1-micro and 8.0\% in F1-macro highlights SADA's enhanced ability to generalize across different domains within this dataset. Similarly, on the LDWBC dataset, SADA scored 86.0\% (F1-micro) and 68.2\% (F1-macro), surpassing the second-best method, SelfReg, by 1.5\% and 0.4\%, respectively. For the Raabin-WBC dataset, SADA achieved F1-micro and F1-macro scores of 87.4\% and 61.9\%, respectively, achieving equally comparable results as Mixup\cite{xu2020adversarial}. On the Zheng-18 dataset, SADA scored 84.3\% (F1-micro) and 62.6\% (F1-macro), outperforming the second-best method, BoDA\cite{umer2023imbalanced}, which achieved scores of 80.5\% and 58.7\%, respectively. The improvements of 3.8\% in F1-micro and 3.9\% in F1-macro showcase SADA's consistent edge in domain invariant feature extraction and generalization. These consistent improvements across diverse datasets underscore SADA's robustness and efficacy in multi-domain blood cell classification.

% For each dataset, our method achieves the best results, except for the Raabin-WBC dataset, where our method performed on par with the Mixup method, with an F1-micro score of 87.4\% and an F1-macro score and 61.9\%. In contrast, our method significantly outperforms the second-best approach on the Acevedo-20 dataset, with F1-micro and F1-macro scores of 79.9\% and 81.6\%, surpassing the second-best method by 9.2\% and 8.0\%, respectively. 

\begin{table}[tb]
  \begin{center}
    \caption{Comparison with the state-of-the-art methods on SYSU3H. The best performance is in \textbf{bold} and the second-best is indicated with \underline{underline}.}
    \label{extenal experiment}
    \begin{tabular}{c|cc} % <-- Alignments: 1st column left, 2nd middle and 3rd right, with vertical lines in between
    \hline
      Methods & F1-micro & F1-macro\\
      \hline
ERM & 82.6$\pm$1.5 & 68.3$\pm$1.3 \\
CORAL & \underline{83.3$\pm$0.6} & \underline{83.3$\pm$0.6} \\
Mixup & 79.8$\pm$2.2 & 79.7$\pm$2.3 \\
SelfReg & 80.8$\pm$1.2 & 80.6$\pm$1.6\\
BoDA & 79.8$\pm$0.9 & 80.0$\pm$0.9 \\
Ours & \textbf{85.4$\pm$0.3} & \textbf{85.1$\pm$0.3} \\
\hline
    \end{tabular}
  \end{center}
\end{table}

\noindent
\textbf{External test.} After training on four public datasets, we also evaluate our method on a private real dataset, collected from the Third Affiliated Hospital of Sun Yat-sen University. Average results from the 3 trials of different compared methods and our method are presented in Table \ref{extenal experiment}. In the external test, SADA demonstrated superior performance compared to other methods, achieving an F1-micro score of 85.4\% and an F1-macro score of 85.1\%. In comparison, the second-best method, CORAL\cite{sun2016deep}, attained scores of 83.3\% in both F1-micro and F1-macro. The improvement of 2.1\% in F1-micro and 1.8\%  in F1-macro demonstrates SADA's robust domain-invariant feature learning capabilities and better handling of class imbalance. The proposed SADA is specifically tailored for the classification of imbalanced blood cell datasets, surpassing other methods and delivering remarkable performance. This result further underscores the practical utility of SADA in real-world scenarios.

\subsection{Visualization Analysis}
\begin{figure*}[tb]
    \centering
    \includegraphics[width=1.0\linewidth]{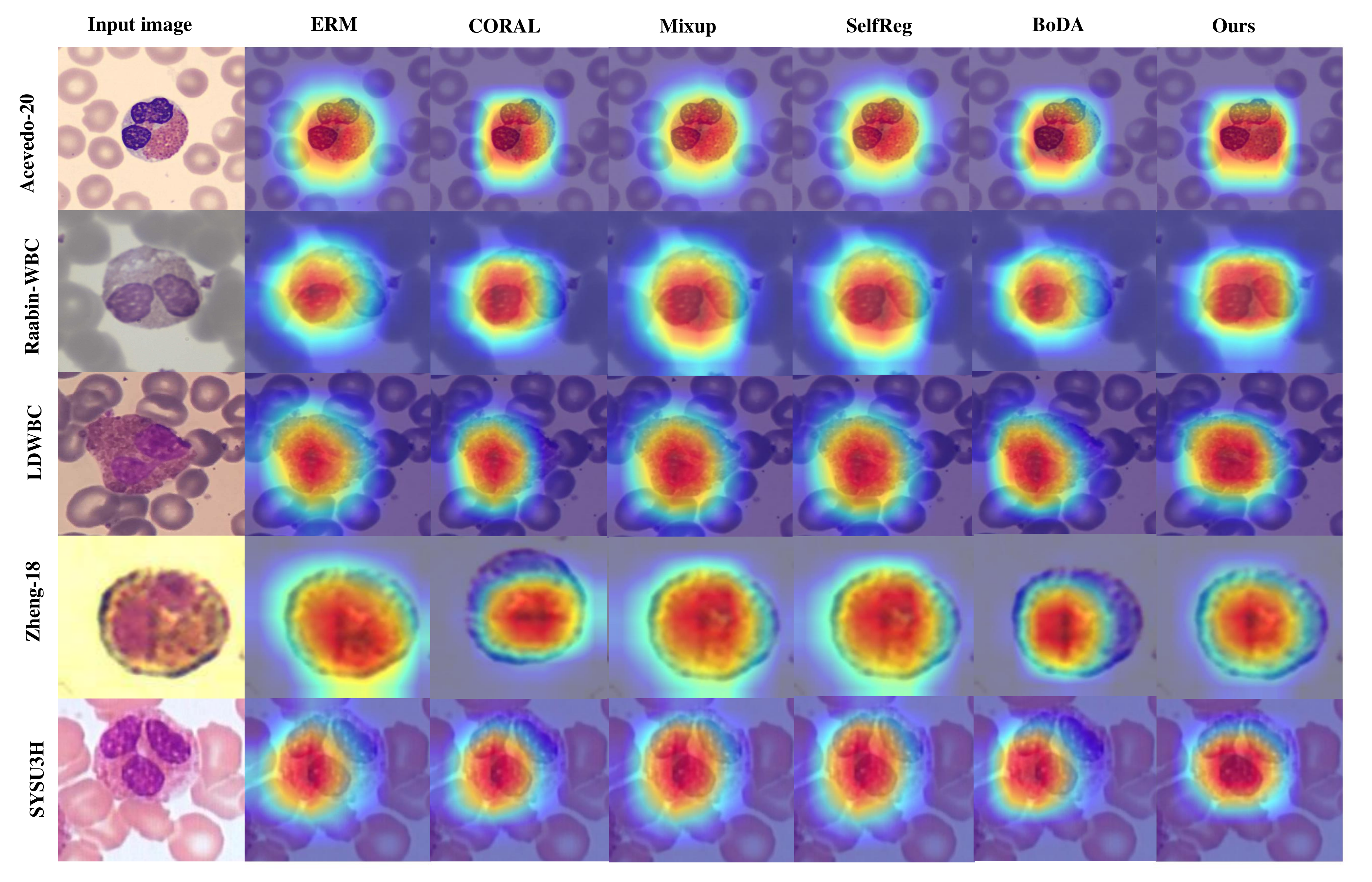}
    \caption{Visualizations of Grad-CAM for ERM, CORAL, Mixup, SelfReg, BoDA, and our proposed method (SADA) on four public datasets and the SYSU3H dataset.}
    \label{fig:gradcam}
\end{figure*} 

% \begin{figure*}[tb]
%     \centering
%     \includegraphics[width=0.98\linewidth]{tsne.pdf}
%     \caption{2D t-SNE visualization of ERM, CORAL, Mixup, SelfReg, BoDA, and our proposed method (SADA) on four public datasets and the SYSU3H dataset.}
%     \label{fig:tsne}
% \end{figure*} 

We employed the Grad-CAM visualization technique to evaluate the effectiveness of our blood cell classification method. As illustrated in Fig. \ref{fig:gradcam}, our method significantly outperforms other approaches, including ERM, CORAL, SelfReg, Mixup, and BoDA. Our approach more effectively targets the blood cell regions, clearly highlighting the critical features. The visual saliency maps generated by our method are more precise and exhibit higher contrast, demonstrating an enhanced ability to extract domain-invariant features. In contrast, the other methods often produce more diffuse and less distinct heatmaps, which can obscure important cellular features and lead to less accurate insights. This improved visual interpretability underscores the value of our method, making it a robust tool for enhancing model transparency and trustworthiness in various blood cell analysis applications.

% We provide visualizations of Grad-CAM for ERM, BoDA, and SADA on four public datasets and the SYSU3H dataset to evaluate the effectiveness of our proposed SADA. As shown in Fig. \ref{fig:gradcam}, our method effectively localizes class-discriminative regions of the images, outperforming others in capturing domain-invariant features.

% We also extract latent representations from the baseline method ERM and our proposed SADA in a leave-one-out setting and visualize them using the t-SNE technique. Fig. \ref{fig:tsne} illustrates that our proposed method better captures domain-invariant features and also demonstrates superior performance in representation learning.

% The 2D t-SNE visualization in Fig. \ref{fig:tsne} demonstrates the efficacy of our method in differentiating between various blood cell types. Our approach produces distinctly clustered representations for each cell type, indicative of superior feature extraction and clear separation of classes. Compared to other methods, our method exhibits more compact and well-defined clusters. The clusters formed by our method are more compact and distinct, indicating a better capture of the underlying data structure and improved feature extraction. This enhanced clustering performance underscores the robustness of our method in distinguishing between various blood cell types, highlighting its potential for precise and reliable blood cell analysis.

\subsection{Hyperparameter Tuning}
\begin{figure*}[tb]
    \centering
    \includegraphics[width=1.07\linewidth]{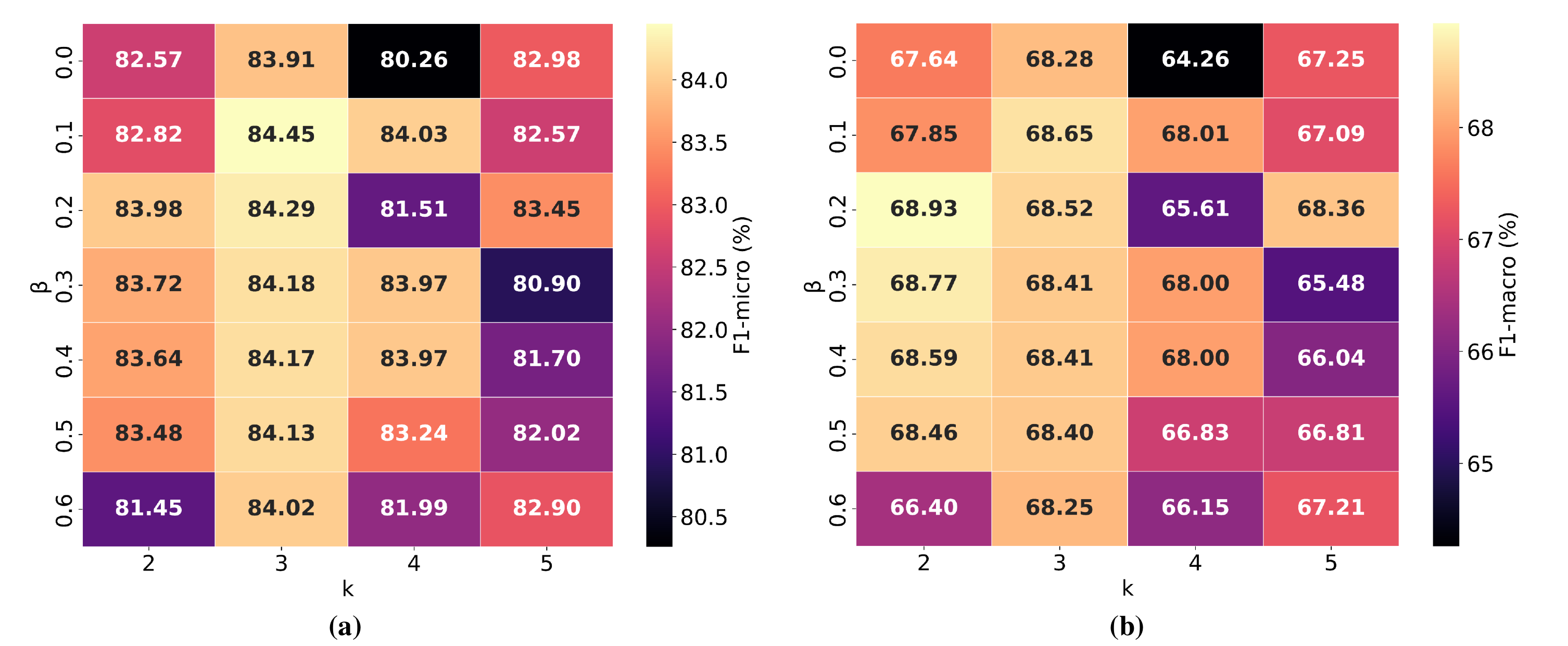}
    \caption{Hyperparameters tuning for SADA. (a) F1-micro score. (b) F1-macro score.}
    \label{fig:hyper}
\end{figure*} 

In this study, we systematically assess the performance of the SADA method using various hyperparameter configurations to determine the optimal settings for our classification task. We experiment with different values for the number of clusters (\(k\)) and the local alignment loss weight parameter (\(\beta\)), evaluating their impact on the F1-micro and F1-macro metrics. This investigation is carried out under a rigorous framework outlined in Section \ref{section 4.1}. The results of hyperparameter tuning averaged across four public datasets, are presented in Fig. \ref{fig:hyper}. We identify \(k = 3\) and \(\beta = 0.1\) as the optimal parameters, achieving an impressive F1-micro score of 84.45\% and a competitive F1-macro score of 68.65\%. These metrics indicate that this parameter set provides superior overall accuracy and balanced class performance compared to all other tested configurations.

\subsection{Ablation Studies}

\begin{table}[tb]
  \begin{center}
    \caption{Ablation study of SADA conducted on four public datasets.}
    \label{ablation}
    \begin{tabular}{cccccc}
    \hline
    
     Variants & $\mathcal{L}_{DISC}$ & $\mathcal{L}_{SC}$ & $\mathcal{L}_{LA}$ &  F1-micro & F1-macro\\
    \hline
    M1 & & & & 79.6$\pm$1.4 & 63.2$\pm$1.5\\
    M2 & \checkmark & & & 83.9$\pm$0.5 & 68.2$\pm$0.8\\
    M3 & & \checkmark & & 78.8$\pm$0.9 & 62.2$\pm$0.7 \\
    Ours & \checkmark & & \checkmark & 84.4$\pm$0.3 & 68.6$\pm$0.2 \\ 
      \hline
    \end{tabular}
  \end{center}
\end{table}

To evaluate the effectiveness of SADA, we conduct ablation experiments on four public datasets with the rigorous setting detailed in Section \ref{section 4.1}. We compare our method with the following three variants: (1) M1 solely employs the ERM algorithm for classification; (2) M2 denotes SADA without local alignment constraint; (3) M3 denotes SADA replacing the domain-invariant supervised contrastive loss with supervised contrastive loss. The experimental results indicate that the absence of the local alignment constraint in M2 results in an F1-micro score of 83.9\% and an F1-macro score of 68.2\%, representing a decrease of 0.5\% and 0.4\%, respectively, compared to the full SADA method. On the other hand, replacing the domain-invariant supervised contrastive loss with the standard supervised contrastive loss in M3 results in significantly lower scores, with an F1-micro of 78.8\% and an F1-macro of 62.2\%, indicating reductions of 5.6\% and 6.4\%, respectively. These results highlight the importance of the domain-invariant supervised contrastive loss in capturing domain-invariant features. The full SADA method achieves the highest performance, with F1-micro and F1-macro scores of 84.4\% and 68.6\%, respectively, validating the synergistic benefits of the proposed components and underscoring its superiority in multi-domain blood cell classification tasks.

% The experimental results for the mentioned three variants are the averages of F1-micro and F1-macro scores on four public datasets. We observe that the absence of the local alignment constraint results in a certain degree of degradation in the model's general performance. Meanwhile, our proposed domain-invariant supervised contrastive learning loss outperforms the supervised contrastive learning loss by a significant margin. 

\section{Conclusion}
\label{con}
In this paper, we propose a novel blood cell classification method called SADA, which consists of stain-based augmentation, local alignment, and domain-invariant supervised learning. The stain-based augmentation effectively enriches sample domain diversity as well as further enhancing the sample volume. With the local alignment constraint, the model can capture domain-invariant features at the feature map level. The domain-invariant supervised learning utilizes the centroid of a raw sample and domain-transformed samples as anchors, enhancing the model's ability to learn discriminative features. The experimental results on four public datasets show that the proposed SADA achieves a new state-of-the-art baseline across all datasets. Specifically, SADA achieved F1-micro and F1-macro scores of 79.9\% and 81.6\% on the Acevedo-20 dataset, 86.0\% and 68.2\% on the LDWBC dataset, 87.4\% and 61.9\% on the Raabin-WBC dataset, 84.3\% and 62.6\% on the Zheng-18 dataset, and 84.4\% and 68.6\% on the average of the four datasets. By achieving the highest scores across multiple datasets and outperforming the second-best method (BoDA) by notable margins, SADA proves to have superior ability to capture domain-invariant features and effectively address class imbalance. Moreover, the proposed SADA achieves F1-micro and F1-macro scores of 85.4\% and 85.1\% on our private real dataset, significantly outperforming the second-best method by 2.1\% in F1-micro and 1.8\% in F1-macro, respectively. This result highlights SADA's potential for application in clinical practice. The consistently higher performance across diverse datasets underscores its robustness and efficacy in multi-domain blood cell classification tasks, marking a significant advancement in the field.

% Extensive experiments on four public blood cell datasets and a private real dataset show that SADA achieves a new state-of-the-art baseline with a big margin. 


\section*{References}
\begin{thebibliography}{10}

\bibitem{wang2021research}
Q.~Wang, T.~Shi, M.~Wan, J.~Wei, F.~Wang, and C.~Mao, ``Research progress of using micro/nanomotors in the detection and therapy of diseases related to the blood environment,'' {\em Journal of Materials Chemistry B}, vol.~9, no.~2, pp.~283--294, 2021.

\bibitem{liu2023revisiting}
Q.~Liu, Z.~Wang, S.~Rong, J.~Li, and Y.~Zhang, ``Revisiting foreground and background separation in weakly-supervised temporal action localization: A clustering-based approach,'' in {\em Proceedings of the IEEE/CVF International Conference on Computer Vision}, pp.~10433--10443, 2023.

\bibitem{van2023pdisconet}
R.~van~der Klis, S.~Alaniz, M.~Mancini, C.~F. Dantas, D.~Ienco, Z.~Akata, and D.~Marcos, ``Pdisconet: Semantically consistent part discovery for fine-grained recognition,'' in {\em Proceedings of the IEEE/CVF International Conference on Computer Vision}, pp.~1866--1876, 2023.

\bibitem{umer2023imbalanced}
R.~M. Umer, A.~Gruber, S.~S. Boushehri, C.~Metak, and C.~Marr, ``Imbalanced domain generalization for robust single cell classification in hematological cytomorphology,'' {\em arXiv preprint arXiv:2303.07771}, 2023.

\bibitem{hehr2023explainable}
M.~Hehr, A.~Sadafi, C.~Matek, P.~Lienemann, C.~Pohlkamp, T.~Haferlach, K.~Spiekermann, and C.~Marr, ``Explainable ai identifies diagnostic cells of genetic aml subtypes,'' {\em PLOS Digital Health}, vol.~2, no.~3, p.~e0000187, 2023.

\bibitem{jamal2020rethinking}
M.~A. Jamal, M.~Brown, M.-H. Yang, L.~Wang, and B.~Gong, ``Rethinking class-balanced methods for long-tailed visual recognition from a domain adaptation perspective,'' in {\em Proceedings of the IEEE/CVF Conference on Computer Vision and Pattern Recognition}, pp.~7610--7619, 2020.

\bibitem{xu2022constructing}
Y.~Xu, Y.-L. Li, J.~Li, and C.~Lu, ``Constructing balance from imbalance for long-tailed image recognition,'' in {\em European Conference on Computer Vision}, pp.~38--56, Springer, 2022.

\bibitem{cai2021ace}
J.~Cai, Y.~Wang, and J.-N. Hwang, ``Ace: Ally complementary experts for solving long-tailed recognition in one-shot,'' in {\em Proceedings of the IEEE/CVF International Conference on Computer Vision}, pp.~112--121, 2021.

\bibitem{li2022nested}
J.~Li, Z.~Tan, J.~Wan, Z.~Lei, and G.~Guo, ``Nested collaborative learning for long-tailed visual recognition,'' in {\em Proceedings of the IEEE/CVF Conference on Computer Vision and Pattern Recognition}, pp.~6949--6958, 2022.

\bibitem{kang2019decoupling}
B.~Kang, S.~Xie, M.~Rohrbach, Z.~Yan, A.~Gordo, J.~Feng, and Y.~Kalantidis, ``Decoupling representation and classifier for long-tailed recognition,'' in {\em International Conference on Learning Representations}, 2019.

\bibitem{Kang_2021}
H.~Kang, T.~Vu, and C.~D. Yoo, ``Learning imbalanced datasets with maximum margin loss,'' in {\em 2021 IEEE International Conference on Image Processing (ICIP)}, IEEE, 2021.

\bibitem{xu2020adversarial}
M.~Xu, J.~Zhang, B.~Ni, T.~Li, C.~Wang, Q.~Tian, and W.~Zhang, ``Adversarial domain adaptation with domain mixup,'' in {\em Proceedings of the AAAI Conference on Artificial Intelligence}, vol.~34, pp.~6502--6509, 2020.

\bibitem{Cubuk_2019}
E.~D. Cubuk, B.~Zoph, D.~Mane, V.~Vasudevan, and Q.~V. Le, ``Autoaugment: Learning augmentation strategies from data,'' in {\em 2019 IEEE/CVF Conference on Computer Vision and Pattern Recognition (CVPR)}, IEEE, 2019.

\bibitem{yao2022pcl}
X.~Yao, Y.~Bai, X.~Zhang, Y.~Zhang, Q.~Sun, R.~Chen, R.~Li, and B.~Yu, ``Pcl: Proxy-based contrastive learning for domain generalization,'' in {\em Proceedings of the IEEE/CVF Conference on Computer Vision and Pattern Recognition}, pp.~7097--7107, 2022.

\bibitem{kim2021selfreg}
D.~Kim, Y.~Yoo, S.~Park, J.~Kim, and J.~Lee, ``Selfreg: Self-supervised contrastive regularization for domain generalization,'' in {\em Proceedings of the IEEE/CVF International Conference on Computer Vision}, pp.~9619--9628, 2021.

\bibitem{cao2019learning}
K.~Cao, C.~Wei, A.~Gaidon, N.~Arechiga, and T.~Ma, ``Learning imbalanced datasets with label-distribution-aware margin loss,'' {\em Advances in Neural Information Processing Systems}, vol.~32, 2019.

\bibitem{lin2017focal}
T.-Y. Lin, P.~Goyal, R.~Girshick, K.~He, and P.~Doll{\'a}r, ``Focal loss for dense object detection,'' in {\em Proceedings of the IEEE International Conference on Computer Vision}, pp.~2980--2988, 2017.

\bibitem{li2020overcoming}
Y.~Li, T.~Wang, B.~Kang, S.~Tang, C.~Wang, J.~Li, and J.~Feng, ``Overcoming classifier imbalance for long-tail object detection with balanced group softmax,'' in {\em Proceedings of the IEEE/CVF Conference on Computer Vision and Pattern Recognition}, pp.~10991--11000, 2020.

\bibitem{zhao2023mdcs}
Q.~Zhao, C.~Jiang, W.~Hu, F.~Zhang, and J.~Liu, ``Mdcs: More diverse experts with consistency self-distillation for long-tailed recognition,'' in {\em Proceedings of the IEEE/CVF International Conference on Computer Vision}, pp.~11597--11608, 2023.

\bibitem{wang2021contrastive}
P.~Wang, K.~Han, X.-S. Wei, L.~Zhang, and L.~Wang, ``Contrastive learning based hybrid networks for long-tailed image classification,'' in {\em Proceedings of the IEEE/CVF Conference on Computer Vision and Pattern Recognition}, pp.~943--952, 2021.

\bibitem{li2021simple}
P.~Li, D.~Li, W.~Li, S.~Gong, Y.~Fu, and T.~M. Hospedales, ``A simple feature augmentation for domain generalization,'' in {\em Proceedings of the IEEE/CVF International Conference on Computer Vision}, pp.~8886--8895, 2021.

\bibitem{nam2021reducing}
H.~Nam, H.~Lee, J.~Park, W.~Yoon, and D.~Yoo, ``Reducing domain gap by reducing style bias,'' in {\em Proceedings of the IEEE/CVF Conference on Computer Vision and Pattern Recognition}, pp.~8690--8699, 2021.

\bibitem{liu2024cross}
Y.~Liu, Y.~Zou, R.~Qiao, F.~Liu, M.~L. Lee, and W.~Hsu, ``Cross-domain feature augmentation for domain generalization,'' {\em arXiv preprint arXiv:2405.08586}, 2024.

\bibitem{chang2021stain}
J.-R. Chang, M.-S. Wu, W.-H. Yu, C.-C. Chen, C.-K. Yang, Y.-Y. Lin, and C.-Y. Yeh, ``Stain mix-up: Unsupervised domain generalization for histopathology images,'' in {\em Medical Image Computing and Computer Assisted Intervention--MICCAI 2021: 24th International Conference, Strasbourg, France, September 27--October 1, 2021, Proceedings, Part III 24}, pp.~117--126, Springer, 2021.

\bibitem{kang2022style}
J.~Kang, S.~Lee, N.~Kim, and S.~Kwak, ``Style neophile: Constantly seeking novel styles for domain generalization,'' in {\em Proceedings of the IEEE/CVF Conference on Computer Vision and Pattern Recognition}, pp.~7130--7140, 2022.

\bibitem{shen2022randstainna}
Y.~Shen, Y.~Luo, D.~Shen, and J.~Ke, ``Randstainna: Learning stain-agnostic features from histology slides by bridging stain augmentation and normalization,'' in {\em International Conference on Medical Image Computing and Computer-Assisted Intervention}, pp.~212--221, Springer, 2022.

\bibitem{tan2024rethinking}
Z.~Tan, X.~Yang, and K.~Huang, ``Rethinking multi-domain generalization with a general learning objective,'' {\em arXiv preprint arXiv:2402.18853}, 2024.

\bibitem{li2018domain}
H.~Li, S.~J. Pan, S.~Wang, and A.~C. Kot, ``Domain generalization with adversarial feature learning,'' in {\em Proceedings of the IEEE Conference on Computer Vision and Pattern Recognition}, pp.~5400--5409, 2018.

\bibitem{khosla2020supervised}
P.~Khosla, P.~Teterwak, C.~Wang, A.~Sarna, Y.~Tian, P.~Isola, A.~Maschinot, C.~Liu, and D.~Krishnan, ``Supervised contrastive learning,'' {\em Advances in Neural Information Processing Systems}, vol.~33, pp.~18661--18673, 2020.

\bibitem{wang2020learning}
S.~Wang, L.~Yu, C.~Li, C.-W. Fu, and P.-A. Heng, ``Learning from extrinsic and intrinsic supervisions for domain generalization,'' in {\em European Conference on Computer Vision}, pp.~159--176, Springer, 2020.

\bibitem{jeon2021feature}
S.~Jeon, K.~Hong, P.~Lee, J.~Lee, and H.~Byun, ``Feature stylization and domain-aware contrastive learning for domain generalization,'' in {\em Proceedings of the 29th ACM International Conference on Multimedia}, pp.~22--31, 2021.

\bibitem{li2021progressive}
L.~Li, K.~Gao, J.~Cao, Z.~Huang, Y.~Weng, X.~Mi, Z.~Yu, X.~Li, and B.~Xia, ``Progressive domain expansion network for single domain generalization,'' in {\em Proceedings of the IEEE/CVF Conference on Computer Vision and Pattern Recognition}, pp.~224--233, 2021.

\bibitem{vahadane2016structure}
A.~Vahadane, T.~Peng, A.~Sethi, S.~Albarqouni, L.~Wang, M.~Baust, K.~Steiger, A.~M. Schlitter, I.~Esposito, and N.~Navab, ``Structure-preserving color normalization and sparse stain separation for histological images,'' {\em IEEE Transactions on Medical Imaging}, vol.~35, no.~8, pp.~1962--1971, 2016.

\bibitem{vapnik1999overview}
V.~N. Vapnik, ``An overview of statistical learning theory,'' {\em IEEE Transactions on Neural Networks}, vol.~10, no.~5, pp.~988--999, 1999.

\bibitem{sun2016deep}
B.~Sun and K.~Saenko, ``Deep coral: Correlation alignment for deep domain adaptation,'' in {\em Computer Vision--ECCV 2016 Workshops: Amsterdam, The Netherlands, October 8-10 and 15-16, 2016, Proceedings, Part III 14}, pp.~443--450, Springer, 2016.

\bibitem{Acevedo_2020}
A.~Acevedo, A.~Merino~Gonz{\'a}lez, E.~S. Alf{\'e}rez~Baquero, {\'A}.~Molina~Borr{\'a}s, L.~Bold{\'u}~Nebot, and J.~Rodellar~Bened{\'e}, ``A dataset of microscopic peripheral blood cell images for development of automatic recognition systems,'' {\em Data in Brief}, vol.~30, p.~105474, 2020.

\bibitem{kouzehkanan2022large}
Z.~M. Kouzehkanan, S.~Saghari, S.~Tavakoli, P.~Rostami, M.~Abaszadeh, F.~Mirzadeh, E.~S. Satlsar, M.~Gheidishahran, F.~Gorgi, S.~Mohammadi, {\em et~al.}, ``A large dataset of white blood cells containing cell locations and types, along with segmented nuclei and cytoplasm,'' {\em Scientific Reports}, vol.~12, no.~1, p.~1123, 2022.

\bibitem{chen2021transmixnet}
H.~Chen, J.~Liu, C.~Hua, Z.~Zuo, J.~Feng, B.~Pang, and D.~Xiao, ``Transmixnet: an attention based double-branch model for white blood cell classification and its training with the fuzzified training data,'' in {\em 2021 IEEE International Conference on Bioinformatics and Biomedicine (BIBM)}, pp.~842--847, IEEE, 2021.

\bibitem{zheng2018fast}
X.~Zheng, Y.~Wang, G.~Wang, and J.~Liu, ``Fast and robust segmentation of white blood cell images by self-supervised learning,'' {\em Micron}, vol.~107, pp.~55--71, 2018.

\bibitem{cha2021swad}
J.~Cha, S.~Chun, K.~Lee, H.-C. Cho, S.~Park, Y.~Lee, and S.~Park, ``Swad: Domain generalization by seeking flat minima,'' {\em Advances in Neural Information Processing Systems}, vol.~34, pp.~22405--22418, 2021.

\end{thebibliography}
\end{document}